# Smart Journey in Istanbul: A Mobile Application in Smart Cities for Traffic Estimation by Harnessing Time Series

Senem Tanberk
Research and Innovation
Huawei Turkey Research and Development Center
Istanbul, Turkey
0000-0003-1668-0365

Mustafa Can
Research and Innovation
Huawei Turkey Research and Development Center
Istanbul, Turkey
mustafa.can2@huawei.com

*Abstract*—In recent decades, mobile applications (apps) have gained enormous popularity. Smart services for smart cities increasingly gain attention. The main goal of the proposed research is to present a new AI-powered mobile application on Istanbul's traffic congestion forecast by using traffic density data. It addresses the research question by using time series approaches (LSTM, Transformer, and XGBoost) based on past data over the traffic load dataset combined with meteorological conditions. Analysis of simulation results on predicted models will be discussed according to performance indicators such as MAPE, MAE, and RMSE. And then, it was observed that the Transformer model made the most accurate traffic prediction. The developed traffic forecasting prototype is expected to be a starting point on future products for a mobile application suitable for citizens' daily use.

## I. Introduction

Traffic congestion is a stressful and time-consuming factor affecting citizens in megacities like Istanbul. It shows cycling characteristics depending on work, holidays, and time. Increasing the population living in the city and high well-being will cause to increase the traffic congestion in Istanbul in a few years. Recently, there is a requirement to design smart infrastructures and tools to assist citizens against the challenges of the traffic problem [12-16].

With the growing popularity of mobile applications (apps), the global mobile application market size rapidly increased. The ubiquitous penetration of smartphones, rising internet usage and speed, and the use of artificial intelligence in mobile applications will cause growth in demand for mobile applications future. Applications for various purposes such as gaming, shopping, mobile health, social networking, and music have been available for many years. Currently, it is possible to integrate IoT, sensors, and AI with apps. So, this led to the development of 'smart city apps'. Smart city apps that use real-time data are estimated to augment exponentially in the coming decade. They are expected to be successful in mitigating problems in the lives of citizens [19].

In this study, we designed a mobile application that will provide accurate and realistic guidance for citizens based on Istanbul districts to easily access traffic prediction for the next hours, and thus support them to plan daily life quickly and effectively. With this AI-powered mobile application, we aimed to provide an optional and effective solution to the time-consuming traffic problem. We named the mobile application as "Smart Journey in Istanbul".

The "Smart Journey in Istanbul" application was launched in Istanbul but is designed and developed to be adaptable to any city in Turkey and all around the world. This is one of those all-inclusive mobile applications that provide a solution for citizens for almost everything including traffic prediction in smart cities. The innovative app isn't designed only for the needs of citizens but also businesses and tourists.

In this project, we are aiming to acquire meaningful data from the traffic and weather dataset of the specific points of Istanbul. With that, congestion times and rush hours will be more visible and more exposed for us to refrain in some of the districts. This study uses more than 14 million traffic data of more than 7 thousand location points and will merge them with more than 50 thousand weather data of the specific district location.

The main contributions of this study can be listed as follows:

- We applied a novel approach for data preparation: Istanbul traffic data provided by Istanbul Metropolitan Municipality is summarized hourly on the basis of 6 main districts and combined with weather-related attributes (temperature, humidity, wind direction, wind speed, and precipitation).

- The combined and summarized dataset is trained with various time series approaches (LSTM, Transformer, and XGBoost) in the literature.

- A new mobile application called Smart Journey in Istanbul was designed and developed, and the simulation of the AI-powered mobile system was tested end-to-end with well-trained models.

- The simulation results were analyzed and the performances of the models were examined. It has been observed that the Transformer model is more successful than other models.

The rest of the paper is structured as follows: Section 2 provides technical information related to methods and metrics. Section 3 presents the proposed approach for traffic prediction, including data preparation, experimental results, as well as mobile application simulation results. And then it evaluates and discusses our findings. Section 4 concludes the article with suggestions for future work.

## II. SYSTEM OVERVIEW

### A. Deep learning/Machine Learning Techniques

*LSTM :* Long short-term memory (LSTM) is an artificial neural network which was introduced to prevent the vanishing gradient problem in traditional RNNs by Hochreiter & Schmidhuber (1997). The LSTM models extend the memory of RNNs so that they may store and learn long-term input dependencies. This memory extension has the ability to recall information for a longer amount of time and so permits reading, writing, and erasing information from their memories.

An LSTM model typically consists of three gates: forget, input, and output gates. The forget gate determines whether existing information will be retained or discarded, the input gate specifies how much new information will be added to the memory, and the output gate determines whether the existing value in the cell contributes to the output. Using the following formulas, the calculations for each gate, input candidate, cell state, and hidden state are performed:

$$\sigma(z) = (1 + e^{-z})^{-1} \quad (1)$$
$$f_t = \sigma(W_f[h_{t-1}, x_t] + b_f) \quad (2)$$
$$i_t = \sigma(W_i[h_{t-1}, x_t] + b_i) \quad (3)$$
$$\tilde{C}_t = tanh(W_c[h_{t-1}, x_t] + b_c) \quad (4)$$
$$C_t = f_t * C_{t-1} + i_t * \tilde{C}_t \quad (5)$$
$$o_t = \sigma(W_o[h_{t-1}, x_t] + b_o) \quad (6)$$
$$h_t = o_t * \tanh(C_t) \quad (7)$$

At time , $x_t$ and $h_t$ correspond to the input and hidden state, respectively, while $f_t, i_t,$ and $o_t$ are the forget, input, and output gates, respectively. $\tilde{C}t$ is the input value to be stored, and the amount of storage is subsequently controlled by an input gate. The symbol of $*$ represents element-by-element multiplication.

The network consists of a convolution layer, two LSTM blocks, and a multilayer perceptron (MLP) head. In the first step, a convolutional layer with 3x3 kernels and rectified linear unit activation functions is followed by two LSTM layers [1] in the LSTM network. The dimension of the LSTM layers' hidden state is 128,64, respectively. In MLP head, the three fully-connected neural network layers follow LSTM blocks. The units of fully-connected layers are, respectively, 128,64 and 1. While the first two dense layers are followed by a layer of ReLU (Rectified Linear Units) [6] activation, the last dense layer is followed by a layer of linear activation. In order to prevent overfitting, L2 (Ridge Regulazation) kernel regularizer with 1e-4 regularization factor [7, 8] is also used to the first completely connected layer.

*Transformer :* A transformer is a model of deep learning that employs the mechanism of self-attention to differentially weight the importance of each part of the input data. Transformers process the full input at once, as opposed to RNNs. The attention mechanism gives context for every point inside the input stream. A single transformer block consists of two significant sublayers: a multi-head attention layer and a position-wise completely connected feed-forward network (FFN). For both sublayers, a residual connection [2] and layer normalization [3] are also employed. Below is a detailed explanation of the two sublayers. A multi-head attention layer is composed of $H$ parallel scaled dot-product attention layers, each referred to as a head [4]. A scaled dot-product attention is a function that transforms a query vector and a collection of key-value pairs into an output vector.

$$Attention(Q, K, V) = softmax\left(\frac{QK^T}{\sqrt{d_k}}\right)V \quad (8)$$

In equation (8), $Q$, $K$, and $V$ represent the matrices stacked by multiple queries, key, and value vectors as rows, whereas $d_k$ represents the dimension of the query/key vectors. Multihead attention goes a step further by first mapping $Q$, $K$, and $V$ into distinct lower-dimensional feature subspaces via distinct dense linear layers, and then calculating attention based on the results.

$$MultiHead(Q, K, V) = Concat(head_1, \dots, head_H)W^o$$
$$where\ head_i = Attention(QW_i^Q, KW_i^K, VW_i^V) \quad (9)$$

The outputs of $H$ heads are then concatenated and projected into a final hidden representation using another dense layer, where $W_i^Q$, $W_i^K$, and $W_i^V$ are the weight matrices of the inner dense layers of each head and Wo is the weight matrices of the outer dense layers.

Similar to LSTM, the transformer model consists of a convolution layer, a transformer encoder block [4], and MLP head. Following a convolutional layer with 3x3 kernels and rectified linear unit [3] activation functions is a transformer encoder block. Each token's embedding size is 256, while the feed-forward network's hidden layer size is 256. Additionally, the transformer block is equipped with four attention heads for the self-attention mechanism. Then, the three fully-connected layers neural networks follow the network. The equivalent units of fully-connected layers are 128 and 64, and 1. Unlike the preceding two dense layers, the final dense layer is followed by a layer of linear activation. To prevent overfitting, L2 (Ridge Regulazation) kernel regularizer with 1e-4 regularization factor [7, 8] is also used to the dense layer.

*XGBoost :* XGBoost is an upgraded machine learning method based on the gradient lifting decision tree that can efficiently construct enhanced trees and operate in parallel. Core to XGBoost is the value of optimizing the objective function. XGBoost makes use of a tree ensemble model consisting of a collection of classification and regression trees (CART) [5]. Using trees as base learners, this sort of boosting is referred to as Tree Boosting. Multiple CARTs can be used in combination, and the final prediction is the total of each CART's score, as a single CART may not be sufficient for achieving satisfactory results. Using the following equations, the model can be expressed as follows:

$$\hat{y}_i = \emptyset(x_i) = \sum_{k=1}^{K} f_k(x_i), f_k \in F, \quad (10)$$

where f is a function in the functional space $F$,

$$F = \{ f(x) = w_q(x) \} (q: \mathbb{R}^m \to T, w \in \mathbb{R}^T) \quad (11)$$

with being the collection of all potential CARTs where $q$ is the structure of each tree that maps an example to the corresponding leaf index. $T$ is the number of leaves on the tree, $w$ is the weight of each leaf, and $K$ is the total number of trees.

$$L^{(t)} = \sum_i^n l(y_i, \hat{y}_i^{(t-i)} + f_t(x_i)) + \Omega(f_t) + C \quad (12)$$

The objective function to be optimized is represented by Equation (12) and is trained additively by adding the $f_t$ function that aids in minimizing the objective. $\hat{y}_i^{(t-i)}$

represents the model prediction of the previous $i$ at iteration $t-1$. $l(y_i, \hat{y}_i^{(t-i)})$ is the training loss function and $\Omega$ represent the regularization function. C is a constant term. The regularization term is used to manage the variance of the fit in order to control the flexibility of the learning task and generate models that generalize to unknown data more effectively. Controlling the model's complexity is necessary to avoid overfitting the training data.

*B. Performance Metrics*

Using the mean absolute percentage error (MAPE), the mean average error (MAE), and the root mean squared error (RMSE) metrics, the error in the estimated total number of vehicles is measured for each estimation method, according to (13)

$$MAPE = \frac{1}{n}\sum_{i=1}^{n}\frac{|Actual_i - Predicted_i|}{Actual_i}$$

$$MAE = \frac{1}{n}\sum_{i=1}^{n}|Actual_i - Predicted_i| \quad (13)$$

$$RMSE = \sqrt{\frac{\sum_{i=1}^{n}(Predicted_i - Actual_i)^2}{n}}$$

## III. EVALUATION

In this project, we conducted extensive experiments via various deep learning or machine learning time series approaches on the Istanbul traffic dataset combined with the weather data. Data preprocessing steps were applied to the dataset. Next, we compared the results of all prediction models. We integrated the trained model with the Smart Journey in Istanbul mobile application developed for this project and tested the resulting system. The artificial intelligence parts of the proposed system were developed with Python. Android and Kotlin were used on the mobile application side. For mobile communication, Flask and REST API were employed.

*A. Datasets*

In this work, we used data in Open Data Portal provided by Istanbul Metropolitan Municipality to retrieve traffic density information hourly based on location in Istanbul [17]. Within the scope of this study, the traffic density information from 06-2020 to 05-2021 is used. To obtain comprehensive and accurate traffic information, we combined weather data from websites in [18].

The retrieved meteorologic data consists of 5 features; temperature, humidity, wind direction, wind speed, and precipitation. It includes the following columns:

- T2M : Temperature at 2 meters
- QV2M : Specific humidity at 2 meters
- WD2M : Wind direction at 2 meters
- WS2M : Wind speed at 2 meters
- PRECTOTCORR : Precipitation

The final dataset for one district (e.g. Tuzla) consists of 7891 traffic density information combined with weather attributes ranging between 06-2020 and 05-2021.

*B. Data Preparation*

IBB (The Istanbul Metropolitan Municipality) shares several open-source datasets in various disciplines including environment, health, and earth science [17]. One of them is the hourly traffic density data set, which contains Istanbul location density and traffic information hourly.

We calculated the traffic difference of certain time zones for only one month's traffic data set in Fig. 1. For this, we plotted and checked the average speed and number of vehicles of check-in and check-out times with the Python Folium library as seen in Fig. 2 and Fig. 3. We only used the first 10.000 data due to Folium's speed restrictions.

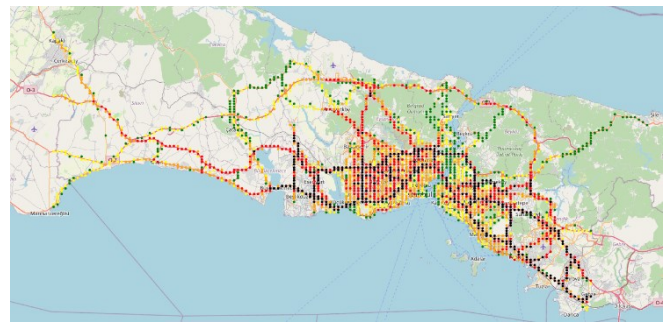

Fig. 1. Data used to create folium plotting in the map.

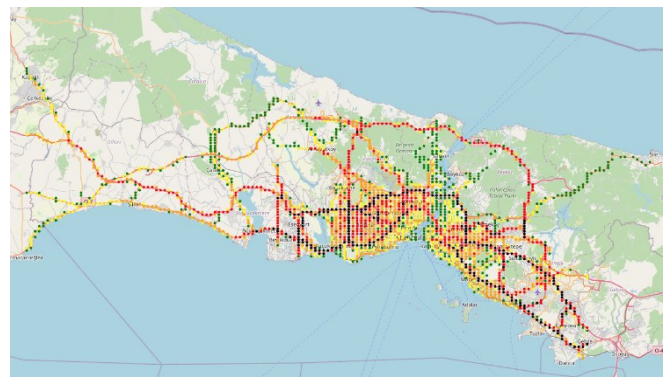

Fig. 2. Traffic congestion of Check-in times (06-09 in the morning).

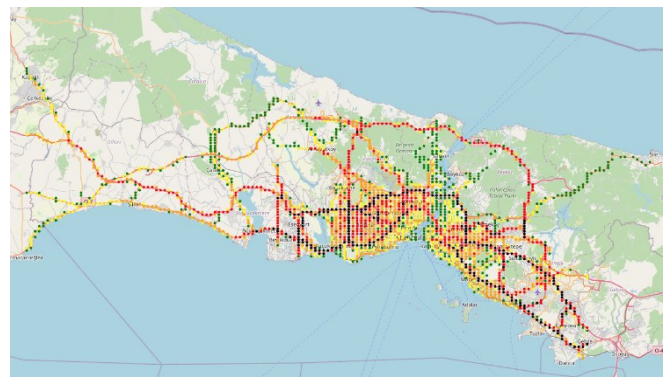

Fig. 3. Traffic congestion of Check-out times (17-20 in the afternoon).

Then we investigated and analyzed the data. We observed that the average speed is not changing while the vehicle numbers go up in the afternoon by %20.

*Morning*
NUMBER_OF_VEHICLES    816233
AVERAGE_SPEED    59.003084

*Afternoon*
NUMBER_OF_VEHICLES    1091036
AVERAGE_SPEED    58.75420489296636

After combining all the traffic data, we had a dataset of 14.183.614 rows and 9 columns. To combine traffic data with the weather dataset in [18], we used the 6 manual points in Fig. 4.

```
[(40.8457, 29.3584, 'TUZLA'),
 (41.0356, 28.8534, 'BAGCILAR'),
 (41.0223, 28.5749, 'BUYUK_CEKMECE'),
 (40.9937, 29.1388, 'ATASEHIR'),
 (41.0822, 28.9862, 'KAGITHANE'),
 (41.0151, 28.9551, 'FATIH')]
```

Fig. 4. Latitude, Longitude, and District Name mapping.

After combining all weather data, we had 57.048 rows and 12 columns in Fig. 5.

Fig. 5. Weather data for 6 districts.

A summary of the vehicle numbers and weather data can be seen in Fig. 6.

Fig. 6. Sum of the vehicle numbers and weather data.

We merged the weather and traffic datasets and created new column called DISTANCE_LOC. This column is calculated closest position of the manually selected 6 districts of Istanbul. The calculation is done by using the Geopy library. Geopy library uses the Harvesine formula for calculating the world distance. In order to analyze the distribution of the number of traffic data in the districts, we used Counter from the collections library of Python in Fig. 7.

```
Counter({'ATASEHIR': 3125813,
         'KAGITHANE': 2487811,
         'TUZLA': 1296943,
         'BAGCILAR': 2939100,
         'BUYUK_CEKMECE': 3522664,
         'FATIH': 811283})
```

Fig. 7. Distribution of traffic data numbers over the districts.

Dataset before preperation and after preperation can be seen in Fig. 8 and Fig. 9 respectively. So the dataset is ready for training by LSTM, Transformers, and XGBoost models for predicting future traffic congestion.

Fig. 8. Data before preparation.

Fig. 9. Data after preparation.

C. *System Architectures*

Figure 10 shows a high-level representation of the Smart Journey in Istanbul software architecture. "Smart Journey in Istanbul" provides a mobile app designed for Android, a popular mobile platform. The application provides an interactive interface that will allow the citizen, in other words user to enter various options such as the district of Istanbul, date, and hour.

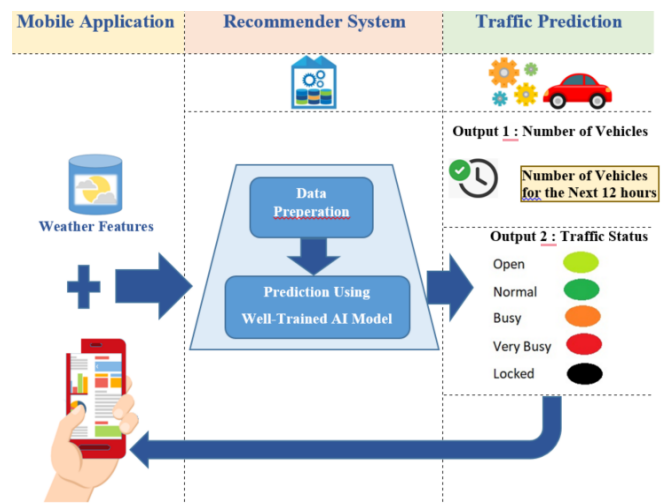

Fig. 10. The mobile application developed in this study: "Smart Journey in Istanbul".

Figure 11 illustrates deep learning or machine learning based traffic prediction system trained with Istanbul traffic data and weather data. Smart Journey in Istanbul mobile app connects to the trained deep learning or machine learning model to apply the traffic prediction. The trained deep learning or machine learning infrastructure provides a data processing pipeline that can extract the traffic prediction based on the training data from the user options provided via the mobile application.

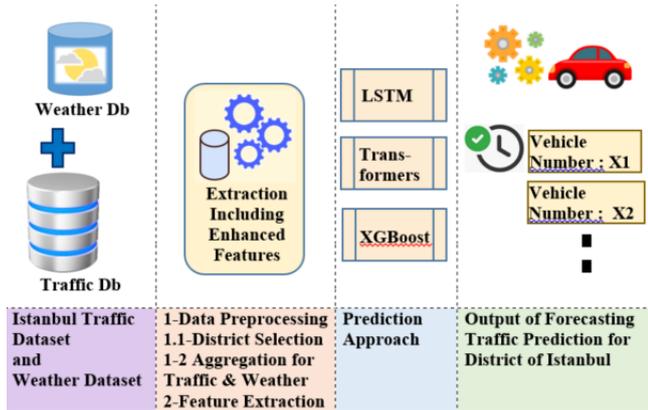

Fig. 11. Recommender system architecture by time series for traffic prediction in Istanbul.

*D. Simulation Results, Analysis and Discussion*

After data preprocessing steps and preparing the data for both training and testing, we applied three prediction time series approaches (LSTM, Transformer, and XGBoost) to the aggregated traffic dataset for six districts of Istanbul. The test results of the algorithms were examined. Comparison results are given in Table 1. Based on MAPE values between 10% and 20% in Table 1, simulation results are good. It has been decided to use the XGBoost model, which gives the optimum results quickly, on the mobile application.

The neural network models' hyperparameters are optimized. For both models, we employed stochastic gradient descent (SGD) [9], with the initial learning rate and momentum fixed at 2.5e-5 and 0.90, respectively. Huber loss function is the cost function of both neural networks [10]. The delta of Huber loss remains constant at 1. In addition, we employed the learning rate scheduler and early stop callback to prevent overfitting and converge neural network models during training. The proposed learning rate scheduler decreases the learning rate by a factor of 0.1 after every 150 epochs. The suggested early halting in the training of both neural networks ends training when the validation loss stops improving. The early termination tolerance is fixed at 5. In addition, neural network implementations were carried out in Python utilizing the Tensorflow framework [11].

In the XGBoost model, the following hyperparameters have been optimized: maximum tree depth, minimum child weight, learning rate (eta), subsample ratio of the training instances, and forest size, which are 5, 4, 0.05, 0.7, 500 respectively. Additionally, we configured early stop round to 15 during training.

TABLE I. EXPERIMENTAL RESULTS

| City | District | Model | MAPE | MAE | RMSE |
|---|---|---|---|---|---|
| Istanbul | FATIH | LSTM | 23% | 0.0597 | 0.0883 |
| | | Transformer | 13% | 0.0515 | 0.0793 |
| | | XGBoost | 15% | 0.0533 | 0.0773 |
| | BUYUK CEK-MECE | LSTM | 14% | 0.0633 | 0.0947 |
| | | Transformer | 13% | 0.0587 | 0.0884 |
| | | XGBoost | 16% | 0.0733 | 0.0982 |
| | ATA-SEHIR | LSTM | 18% | 0.0581 | 0.0865 |
| | | Transformer | 12% | 0.0506 | 0.0774 |
| | | XGBoost | 13% | 0.0582 | 0.0862 |
| | KAGIT-HANE | LSTM | 12% | 0.0633 | 0.0951 |
| | | Transformer | 10% | 0.0538 | 0.0861 |
| | | XGBoost | 15% | 0.0749 | 0.0990 |
| | TUZLA | LSTM | 12% | 0.0441 | 0.0655 |
| | | Transformer | 10% | 0.0373 | 0.0590 |
| | | XGBoost | 11% | 0.0423 | 0.0626 |
| | BAG-CILAR | LSTM | 14% | 0.0610 | 0.0902 |
| | | Transformer | 11% | 0.0517 | 0.0815 |
| | | XGBoost | 13% | 0.0578 | 0.0835 |

In this study, the main aim was to predict city density for Istanbul districts for the next 12 hours, through the mobile application whose user interface is shown in Figure 12. Experimental studies can be expanded by adding various features of events in the city (congress, rally, match, concert, etc.) to the traffic data set. So, the system can actively guide with artificial intelligence, taking into account the current conditions.

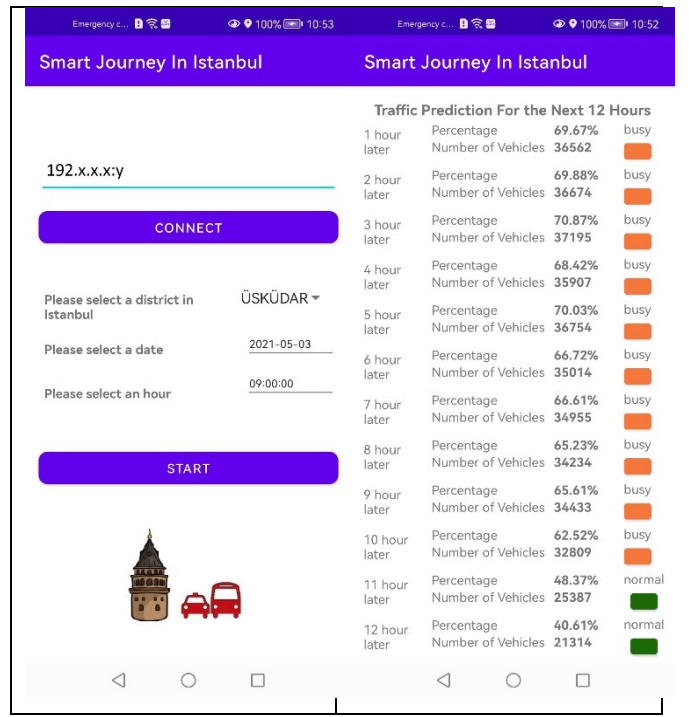

Fig. 12. Mobile application screen for "Smart Journey in Istanbul".

IV. CONCLUSIONS

In this study, we presented a new AI-powered mobile application based on time series to predict traffic for citizens in Istanbul. The well-known time series models (LSTM, Transformer, and XGBoost) were trained with a traffic dataset enriched with weather data and successfully predicted traffic density. The trained AI model was integrated into the mobile application called Smart Journey in Istanbul, and the proposed system was tested end-to-end in the mobile test environment.

In the future, different insights could be explored. First, the time series forecasting could be reconsidered for long-term using transformers auto-encoders. In addition, online updating of traffic data and weather data is expected. It is aimed to enrich the traffic data with various features of the current events (congress, rally, match, concert, etc.) in the city.